\documentclass[pmlr]{jmlr}

\usepackage{booktabs}
\usepackage[load-configurations=version-1]{siunitx} 

\theorembodyfont{\upshape}
\theoremheaderfont{\scshape}
\theorempostheader{:}
\theoremsep{\newline}

\jmlrvolume{86}
\jmlryear{2018}
\jmlrsubmitted{submission date}
\jmlrpublished{publication date}
\jmlrworkshop{2nd IJCAI Workshop on Artificial Intelligence in Affective Computing} 

\title[Individualized Affective Experience Decoding]{Facial Expression and Peripheral Physiology Fusion\\ to Decode Individualized Affective Experience}

  \author{\Name{Yu Yin, Mohsen Nabian, Sarah Ostadabbas\nametag{\thanks{This paper has code available at GitHub under \href{https://github.com/ostadabbas/3d-facial-landmark-detection-and-tracking}{3D Facial Landmark Detection and Tracking}, provided by the corresponding author's lab.}}} \Email{yin.yu1@husky.neu.edu, monabiyan@ece.neu.edu, ostadabbas@ece.neu.edu}\\
  \Name{Miolin Fan, ChunAn Chou} \Email{fan.mi@husky.neu.edu, ch.chou@northeastern.edu}\\
  \Name{Maria Gendron} \Email{maria.gendron@gmail.com}\\
   \addr Northeastern University, Boston, USA}

\usepackage{times}
\usepackage{xcolor}
\usepackage{soul}
\usepackage[utf8]{inputenc}
\usepackage[small]{caption}
\usepackage{graphicx}
\usepackage{amsmath}
\usepackage{amssymb}

\newcommand{\figref}[1]{Fig.~\ref{fig:#1}}
\newcommand{\tblref}[1]{Table~\ref{tbl:#1}}

\begin{document}

\newcommand{\rating}{
\begin{figure}[t]
 \centering
 \includegraphics[width=0.55\linewidth,trim={0in 0in 0in 0in}, clip=true]{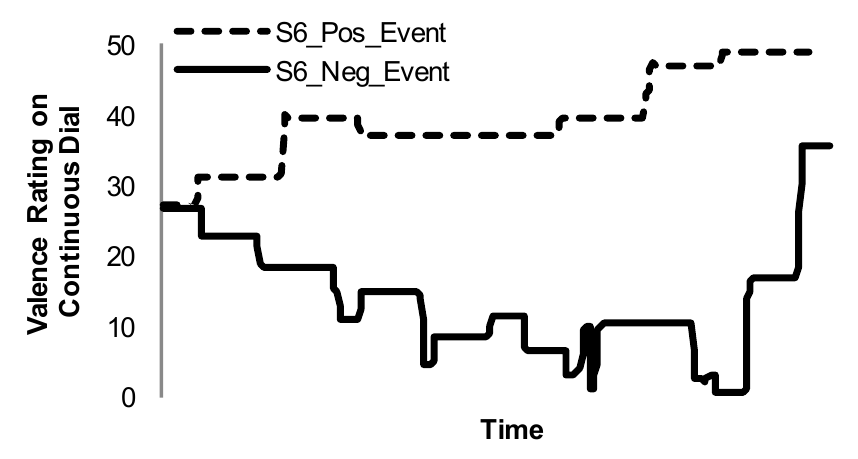}
 \caption{Continuous rating of an example video stimulus using rating dial (ranging from negative = 0 to positive = 50) reveals dynamics in videos across time.
}
\vspace{-.2in}
\label{fig:rating}
\end{figure}
}

\newcommand{\landmarkFramework}{
\begin{figure}[t]
 \centering
 \includegraphics[width=0.9\linewidth,trim={0in 0in 0in 0in}, clip=true]{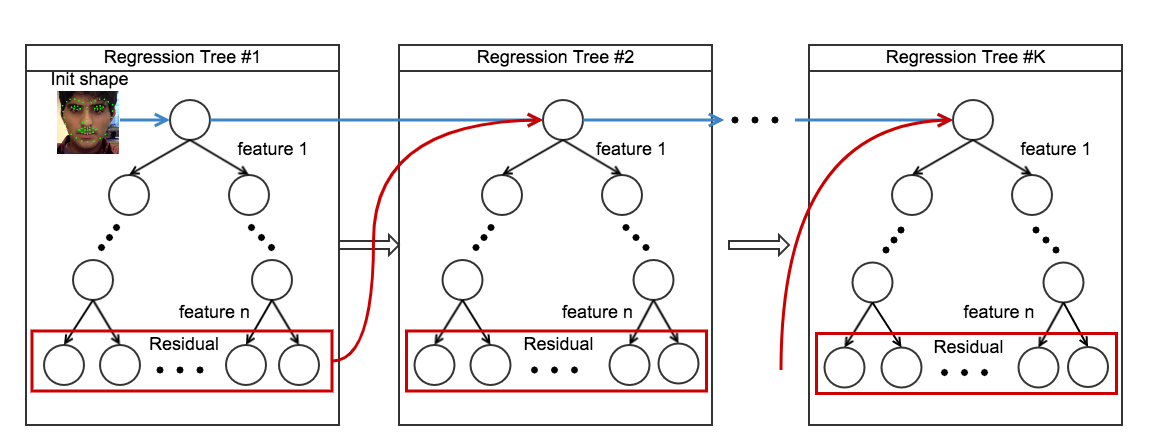}
 \caption{The framework of the cascade of regression trees designed for 2D facial landmark detection. In each level of cascade, estimated landmarks are refined by adding residuals produced by the previous regression tree.}
 \vspace{-.2in}
\label{fig:landmarkFramework}
\end{figure}
}

\newcommand{\MohsenLandmark}{
\begin{figure}[h]
 \centering
 \includegraphics[width=0.45\linewidth,trim={0in 0in 0in 0in}, clip=true]{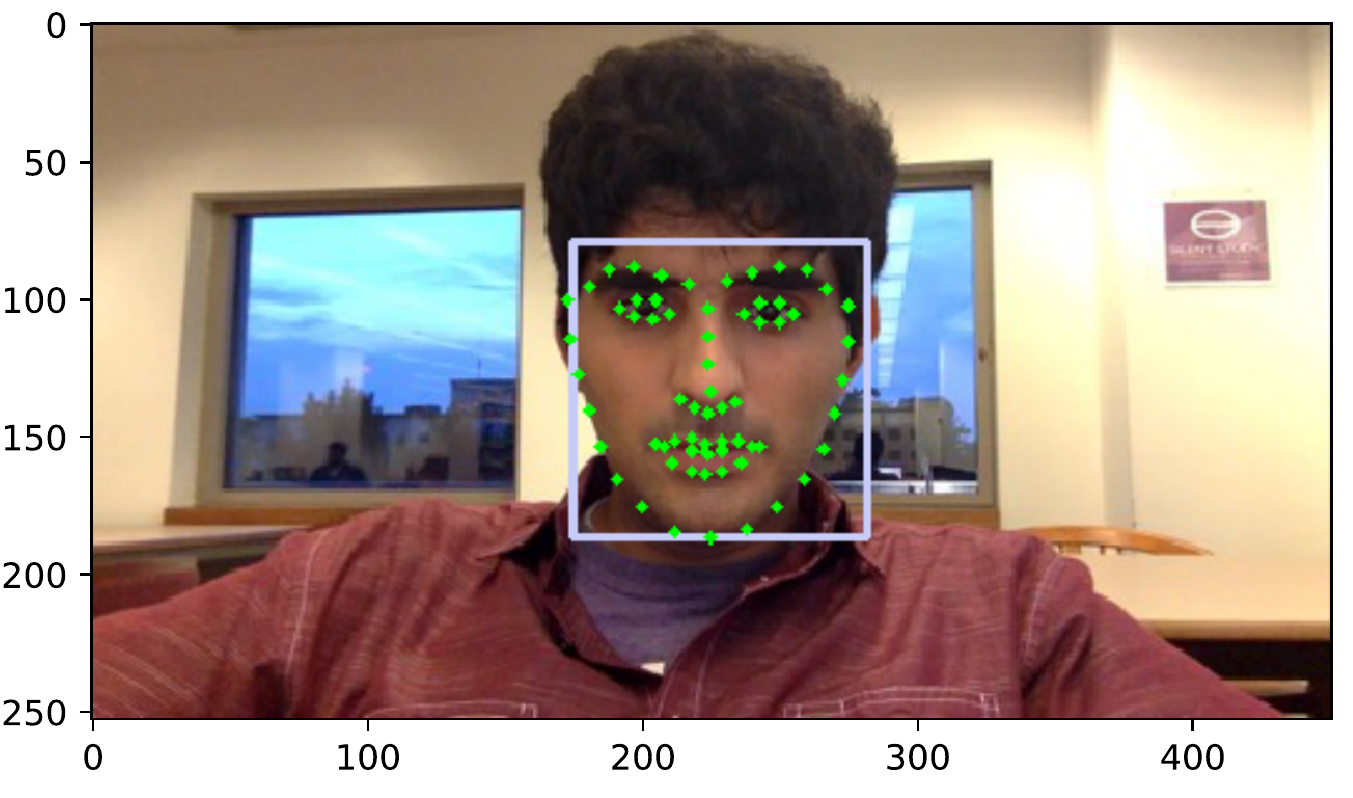}
 \caption{Results of 2D landmark detection on a webcam video frame. It outputs 68 landmarks inside a bounding box detected using HOG features.}

\label{fig:MohsenLandmark}
\end{figure}
}

\newcommand{\YuModel}{
\begin{figure}[t]
 \centering
  \begin{tabular}{@{}cc@{}}
  \multicolumn{1}{c}{\includegraphics[width=0.28\linewidth]{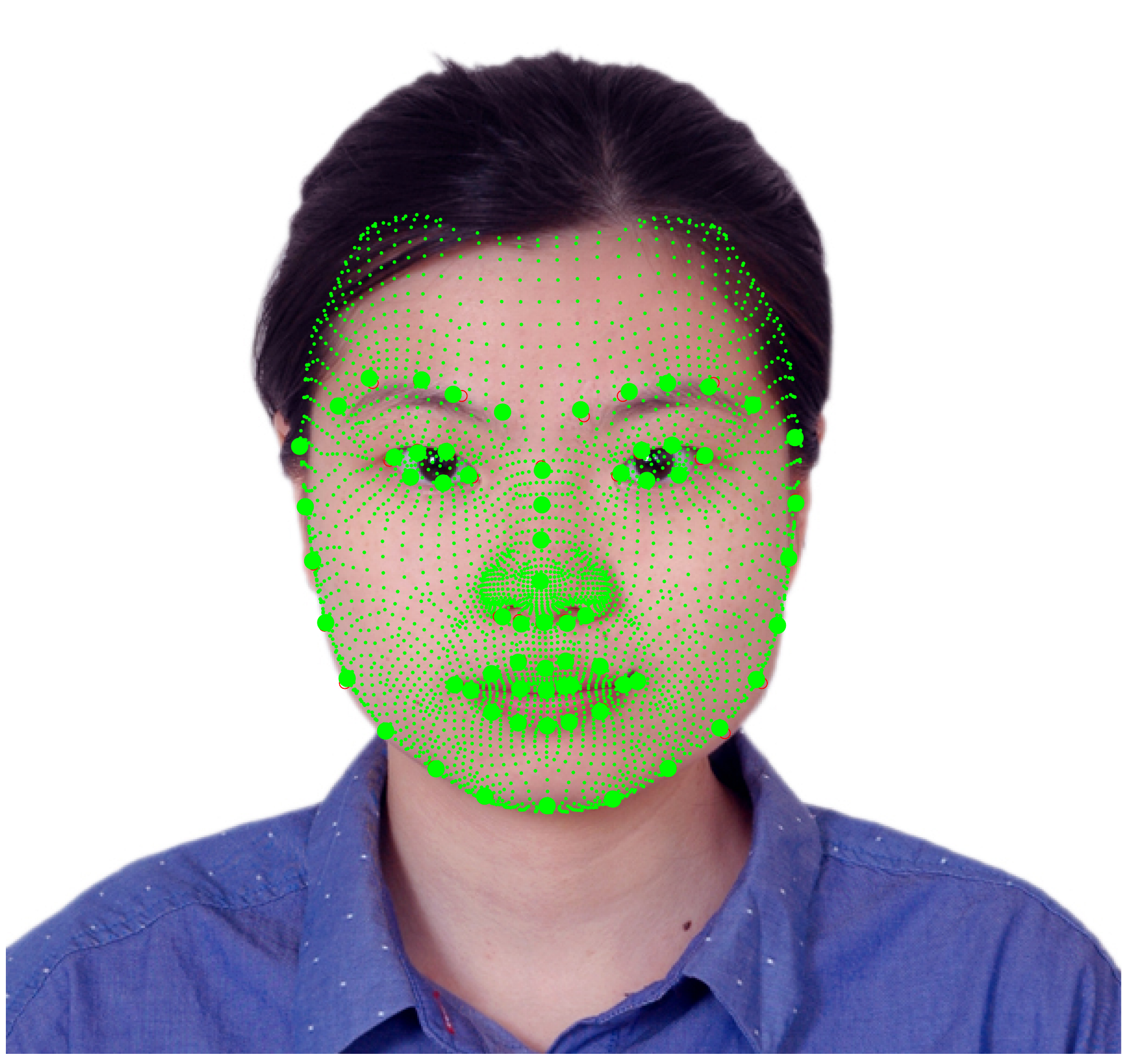}}
 \includegraphics[width=0.4\linewidth,trim={0in 0in 0in 0in}, clip=true]{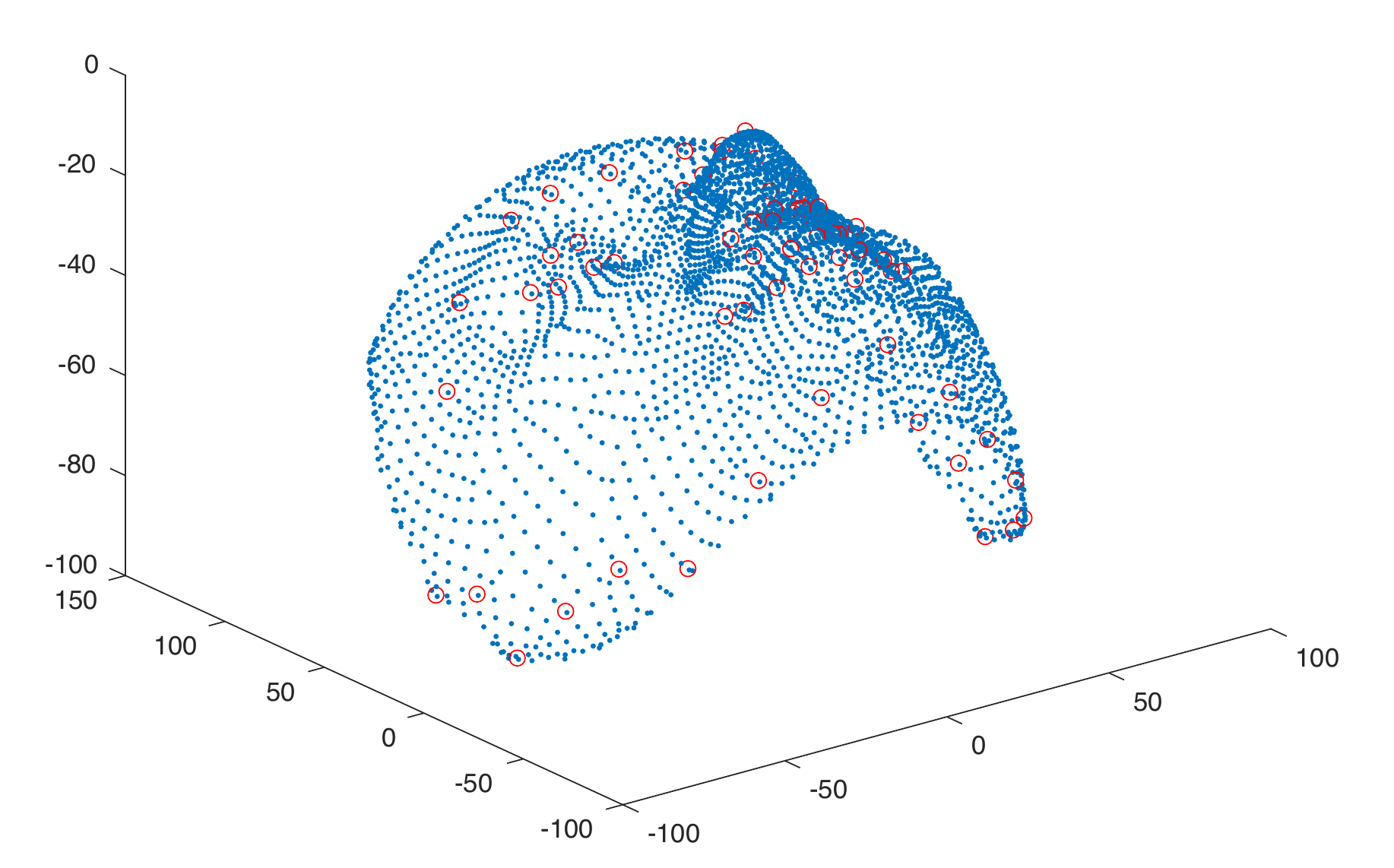}&
 \includegraphics[width=0.28\linewidth,trim={0in 0in 0in 0in}, clip=true]{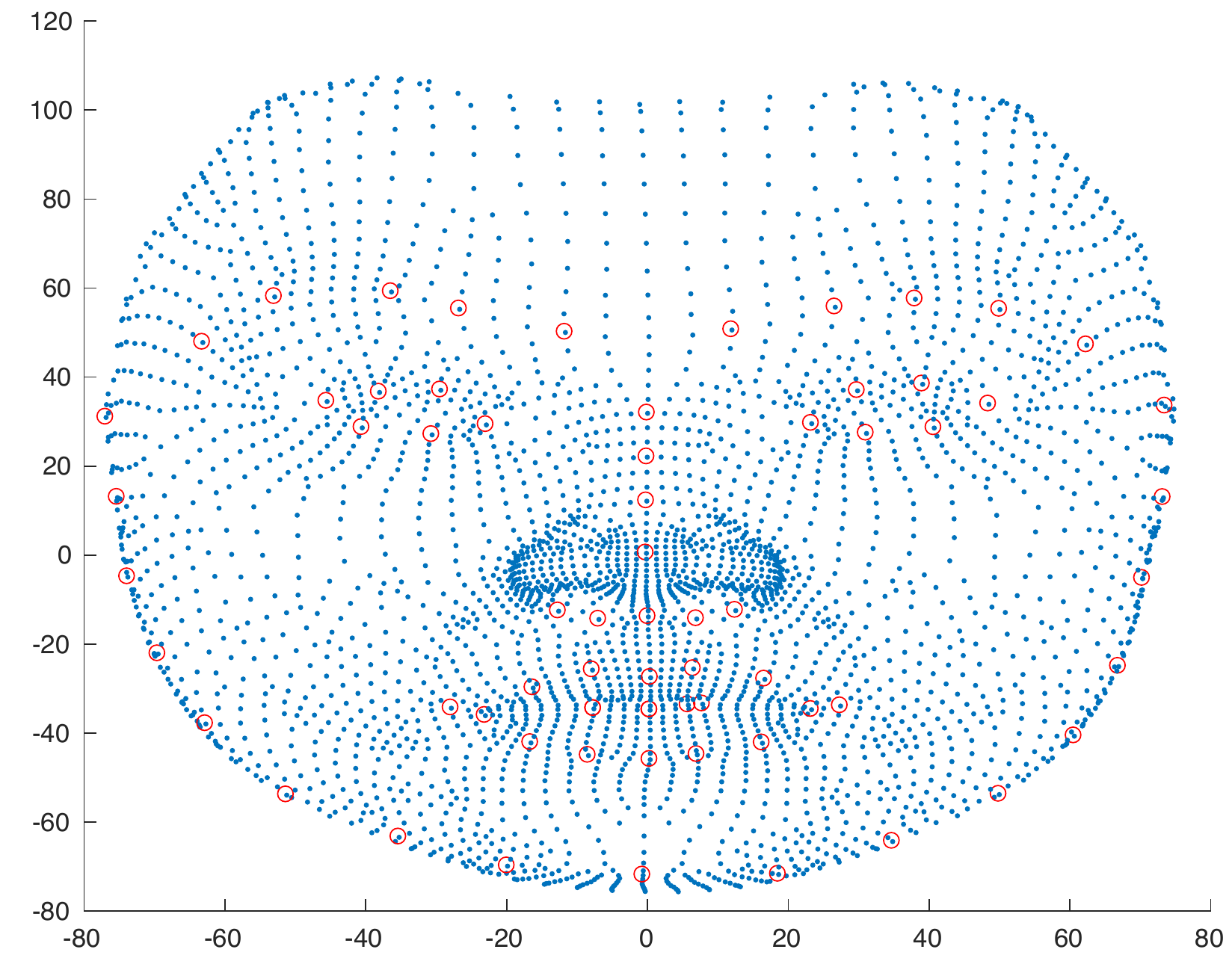}
  \end{tabular}
 
 \caption{An example result of the landmark fitting: resulting shape and model fitting (left), 3D model fitting result (middle), and frontal view of the 3D facial model (right).}
 \vspace{-.2in}
\label{fig:YuModel}
\end{figure}
}

\newcommand{\graphFlow}{
\begin{figure*}[t]
 \centering
 \includegraphics[width=1\linewidth,trim={0in 2.5in 0in 1.5in}, clip=true]{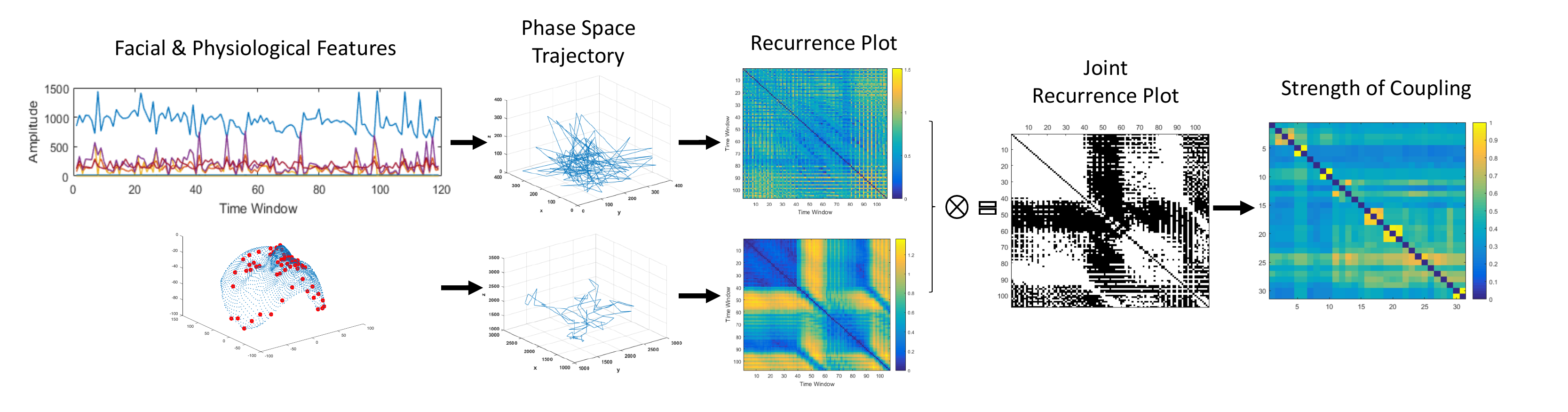}
 \caption{The framework for multimodal data fusion using recurrent network. First, signal-specific facial and physiological features are windowed and extracted. Second, the phase space trajectory for each modality is reconstructed and the corresponding recurrence plot is obtained. Then, a joint recurrence plot (JRP) is calculated by multiplying recurrence plots together. Finally, complex network metrics are extracted to assess the inter-system dynamical coupling.}
 \vspace{-.2in}
\label{fig:graphFlow}
\end{figure*}
}

\newcommand{\lndNo}{
\begin{figure}[ht]
 \centering
 \includegraphics[width=1\linewidth,trim={0in 2in 0in 1in}, clip=true]{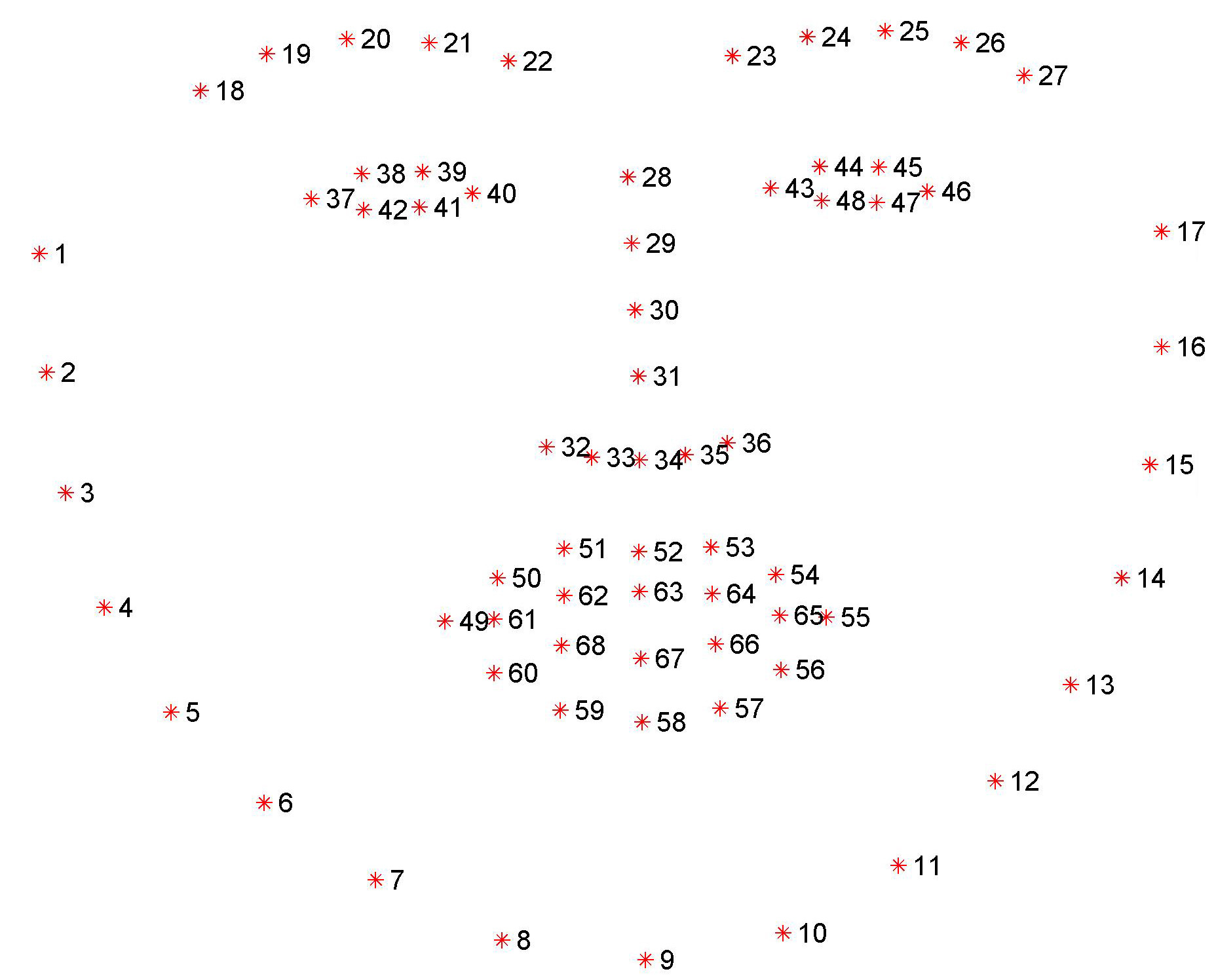}
 \caption{The framework for feature fusion.}
 \vspace{-.2in}
\label{fig:lndNo}
\end{figure}
}

\newcommand{\figSCR}{
\begin{figure}
  \centering
  \includegraphics[width=\linewidth,trim=0in 0in 0in 0in, clip=true]{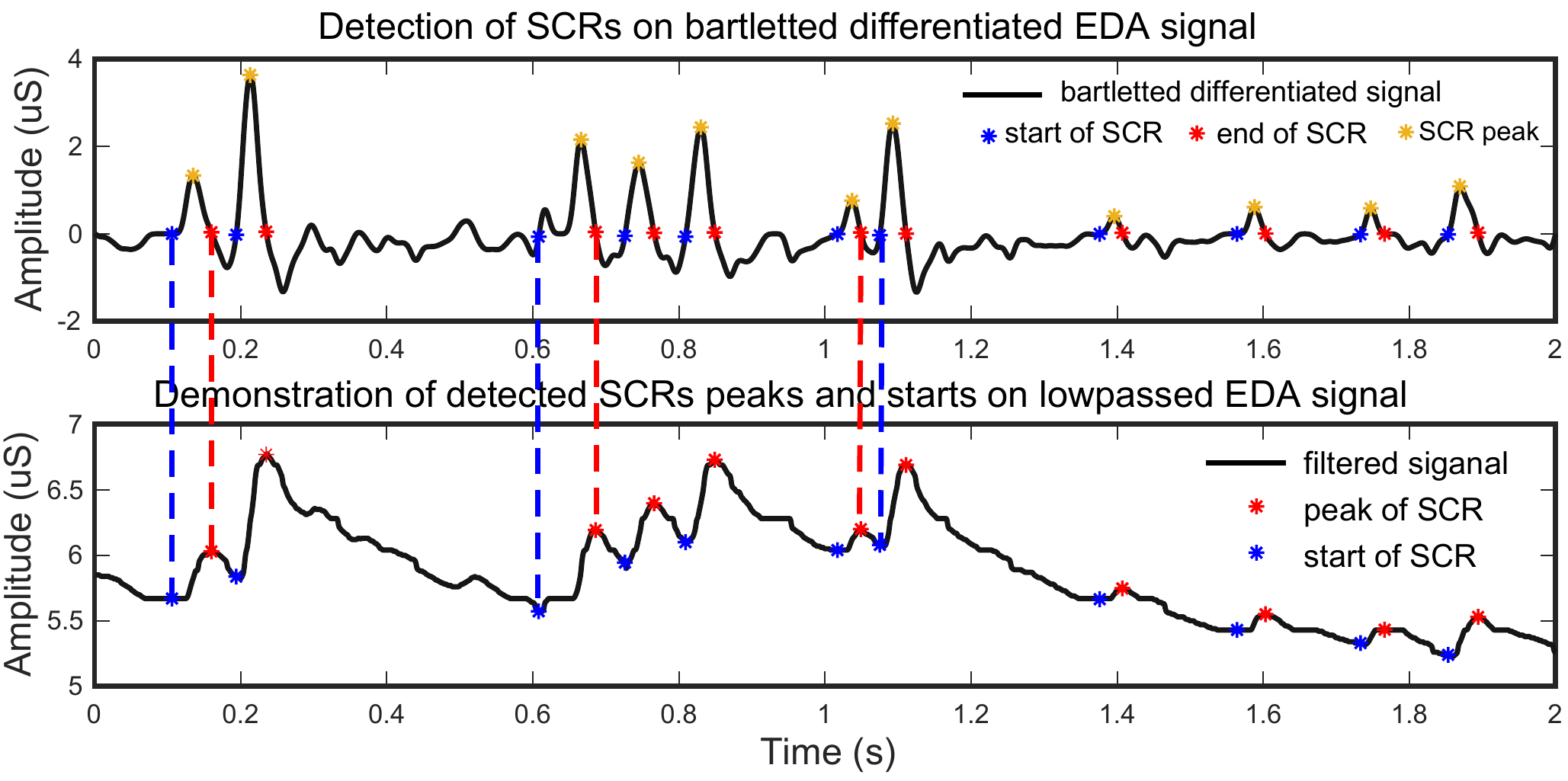}
  \caption{Detection of SCRs on bartletted differentiated EDA signal (top) and demonstration of detected SCRs' peaks and starts on low-passed EDA signal (bottom).}
\label{fig:SCR}
\vspace{-.2in}
\end{figure}
}

\newcommand{\figECGwithQRS}{
\begin{figure}[ht]
  \centering
  \includegraphics[width=\linewidth,trim=0in 0in 0in 0in, clip=true]{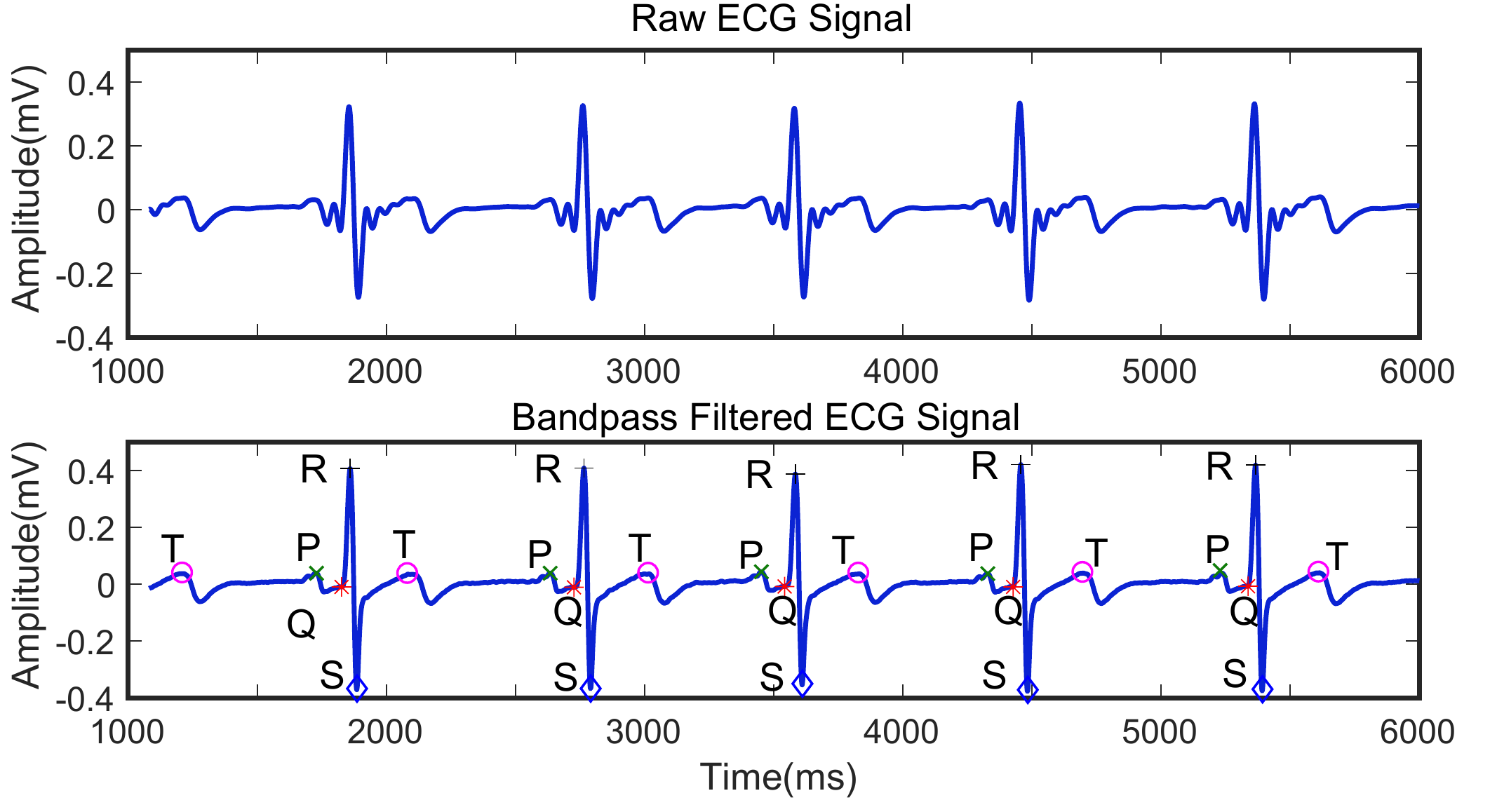}
  \caption{Raw ECG signal (top); bandpass filtered ECG signal (bottom) with the characteristic points (P,Q,R,S and T) detected on the signal.}
\label{fig:ECGwithQRS}
\vspace{-.2in}
\end{figure}
}
\newcommand{\tabFeatureExtraction}{
\begin{table}[h]
\caption{Signal-specific feature extraction parameters.}
\vspace{-.2in}
\begin{center}
  \begin{tabular}{c | c  c  c } 
     & Window Size & Overlap  & Features \\
     Modality& (sec) & (\%)  & (\#) \\
\hline

\textbf{Face} & 5  & 50 & 6 \\

\textbf{Head} & 5  & 50 & 6 \\

\textbf{ECG} & 5  & 50 & 10 \\

\textbf{EDA} & 20 & 88.2 & 5 \\

\textbf{Resp} & 30  & 92.4 & 4
\end{tabular}
\label{tbl:FeatureExtraction}
 \vspace{-.2in}
\end{center}
\end{table}
}

\newcommand{\tabBioFeatures}{
\begin{table}[ht]
\caption{Physiology-specific feature extraction from physiological signal modalities.}
 \vspace{-.2in}
\begin{center}
  \begin{tabular}{ p{1.4cm} p{12cm}}
  \hline
  \hline
    Modality & Extracted Features\\
    \hline
    \textbf{ECG} & Features based on QRS detection: mean R-R intervals (the time between consecutive heartbeats), standard deviation of R-R intervals, standard deviation of the differences between adjacent R-R intervals, the square root of the mean of the sum of the squares of differences between adjacent R-R intervals, the number of pairs of adjacent R-R intervals where the first R-R interval exceeds the second R-R interval by more than 50ms, the number of pairs of adjacent R-R intervals where the second R-R interval exceeds the first R-R interval by more than 50ms, mean area of each QRS complex and its standard deviation. \\
      \hline
    \textbf{EDA} &  Signal mean, numbers of detected SCRs, mean SCR duration, mean SCR amplitude, mean SCR rise-time (where rise-time of an SCR is defined as the time between the initial rise and the peak of an SCR). \\
      \hline
    \textbf{Resp} &  Respiration rate (peak to peak in ms), amplitude (height of peak), percent inhalation (the proportion of rising part of the signal in each cycle) and percent exhalation (the proportion of falling part of the signal in each cycle).\\
    
  \hline
  \hline
\end{tabular}
\label{tbl:BioFeatures}
 \vspace{-.2in}
\end{center}
\end{table}
}

\newcommand{\tabclassificationResult}{
\begin{table}[b]
\caption{Emotional video-based stimulus content prediction results.}
\vspace{-.2in}
\begin{center}
  \begin{tabular}{c | c  c | c  c  c} 
    Modality  & Accuracy & p-value & F1  & Precision & Recall \\
\hline

\textbf{ECG} & 51.4 &0.29& 51.3 & 51.3 & 52.1\\

\textbf{EDA} & 48.6 &0.18& 47.7 & 48.6 & 47.2\\

\textbf{Resp} & 49.3 &0.20& 43.4 & 49.1 & 40.3\\

\textbf{Face} & 45.1 &0.10& 44.6 & 44.9 & 44.4\\

\textbf{Head} & 50.7 &0.27& 52.4 & 50.6 & \textbf{55.6}\\
  \hline

\textbf{Fusion} & \textbf{55.3}&0.49 & \textbf{54.1} & \textbf{56.7} & 51.8\\

\textbf{Random} &  50.4 &0.24& 50 & 50.6 & 48.5

\end{tabular}
\label{tbl:classificationResult}
 \vspace{-.2in}
\end{center}
\end{table}
}

\newcommand{\tabregressionResult}{
\begin{table}[t]
\caption{Affective self-rating scale prediction results.}
\vspace{-.2in}
\begin{center}
  \begin{tabular}{c | c  c } 
    Modality & RMSE & MAE \\
\hline

\textbf{ECG} & 0.33 & 0.28 \\

\textbf{EDA} & 0.27 & 0.24 \\

\textbf{Resp} & 0.27 & 0.24 \\

\textbf{Face} & 0.32 & 0.27 \\

\textbf{Head} & 0.45 & 0.33 \\

\hline

\textbf{Facial} & 0.27 & \textbf{0.24} \\

\textbf{Physio} & 0.29 & 0.25 \\

\textbf{Fusion} & \textbf{0.26} & \textbf{0.24} \\

\textbf{Random} &  0.40 & 0.37
\end{tabular}
\label{tbl:regressionResult}
 \vspace{-.2in}
\end{center}
\end{table}
}

\maketitle

\begin{abstract}
Affective experience prediction using different data modalities measured from an individual such as their facial expression or physiological signals has received substantial research attention in recent years. However, most studies ignore the fact that people besides having different responses under affective stimuli, may also have different resting dynamics (embedded in both facial and physiological patterns) to begin with. In this paper, we present a multimodal approach to simultaneously analyze facial movements and several peripheral physiological signals to decode individualized affective experiences under positive and negative emotional contexts, while considering their personalized resting dynamics. We propose a person-specific recurrence network to quantify the dynamics present in the person's facial movements and physiological data. Facial movement is represented using a robust head vs. 3D face landmark localization and tracking approach, and physiological data are processed by extracting known attributes related to the underlying affective experience. The dynamical coupling between different input modalities is then assessed through the extraction of several complex recurrent network metrics. Inference models are then trained using these metrics as features to predict individual's affective experience in a given context, after their resting dynamics are excluded from their response. We validated our approach using a multimodal dataset consists of (i) facial videos and (ii) several peripheral physiological signals, synchronously recorded from 12 participants while watching 4 emotion-eliciting video-based stimuli. The affective experience prediction results signified that our multimodal fusion method improves the prediction accuracy up to 19\% when compared to the prediction using only one or a subset of the input modalities. Furthermore, we gained prediction improvement for affective experience by considering the effect of individualized resting dynamics. 
\end{abstract}

\section{Introduction}
Affective experience is an important construct in explaining several critical aspects of human behaviors and is impaired or irregular in a number of neurodevelopmental and psychiatric disorders \citep{panksepp2004affective}. Experimental evidence indicates that positive or negative affective experience plays an important role in motivating future actions \citep{Barrett2009} and can promote behavioral patterns linked to compromised mental health \citep{Wichers2015}. In addition, information about affective states of users has become more and more important in human-computer interaction and many other emerging areas \citep{Schaaff2009} in recent years as it greatly facilitates the ability of computers to heed the rules of human communication \citep{Picard2000}.



A common approach to quantify the range of human affective states and to predict human affective experiences under different circumstances is based on decoding their facial movements. Movements of the face are considered as a particularly rich source for affective display and are commonly referred to as ``facial expressions'' \citep{Littlejohn2008, Ekman2002,facialSynchroney,duran2017coherence,russell2017science,fernandez2013emotion}. However, not all emotions occur with an expression in face or even distinguishable facial expression \citep{Ekman1993} and the correspondence between specific facial expressions and underlying emotional experiences is not robust in psychology \citep{Reisenzein2013}. To address this problem, various physiological signals including electrocardiogram (ECG), electroencephalogram (EEG), electrodermal activity (EDA), blood pressure, and respiration patterns have been used as complementary information to decode affective states \citep{Verma2014,khalaf2017analysis} based on the function of multiple physiological systems in the body \citep{Liao2006,perez2017decoding}.


Among humans, there are considerable individualized differences in facial expression and peripheral physiological responses under similar affective experiences, which influence the recognition results of generalized predictive algorithms. Therefore, There is a large body of literature investigating personalized models and their impact on accurately predicting person-specific affect \citep{Yang2014, Hernandez2011, Kandemir2014}. However, it is crucial to note that human facial movements or physiological responses even without any emotional stimulus (i.e. under resting state) differ substantially from one individual to another, due to multiple intrinsic and extrinsic factors (e.g. gender, personality, cultural influences, level of education, etc.) \citep{Hurlburt2015, Krys2016, Joo2012}. In the majority of the affective computing or even psychological studies, individual baseline differences are only accounted by subtracting the mean value (and very rarely considering the standard deviation) of the collected data during resting state from the stimulus-responding state data, assuming that the resting dynamics can simply be modeled solely with the zeroth and first moments.

In this paper, we applied recurrence network analysis for multimodal data (i.e. facial movements and physiological signals) fusion to identify and decode individualized affective experiences. To extract features from facial movements, we developed a robust landmark tracking approach, in which head movement is also independently tracked and decoupled from the facial landmark movements. As for the physiological signals including ECG, EDA, and respiration, a series of signal-specific algorithms developed in \citep{NabianMohsen2017Abpt, bioSp} were utilized to extract psychologically-related features. We used complex network metrics to assess the inter-system dynamical coupling between different response 
modalities of a person under a negative or positive affective experience. We used these metrics to build an inference model for individual affective experience prediction. Critically, we also modeled the resting dynamics of each individual participant before undergoing any emotional induction, to account for individualized baseline differences in affective experience. Our main contributions in this paper are as follows: (1) employing a 3D model for facial landmark localization/tracking, which decouples head motion from face expression; (2) assessing the resting/affective multimodal response of each individual through a higher order dynamics using recurrence network metrics; and (3) developing a novel multimodal feature fusion approach based on recurrence network for affective experience decoding. 

\section{Related Work}
Emotion recognition studies based on facial expression analysis often take a categorical approach, where a label from a set of six purported basic emotions (anger, disgust, fear, happiness, sadness, surprise) is assigned  to a pattern of facial movements \citep{russell1994there}. Yet in real life, emotions are much more complex \citep{Barrett2016}, and specificity and consistency of facial movements to emotions is often lacking \citep{fernandez2013emotion, Reisenzein2013}. Moreover, some of the emotions do not even fit well in any of the basic categories \citep{Koelstra2013}. 
A finer-grained assessment of facial expressions is to directly detect specific facial muscle actions (action units; AUs), including but not limited to the facial movements on which the basic emotion expressions were based \citep{tian2001recognizing,Cohn2007,sanchez2016cascaded}. The facial features used in AUs recognition studies are often either geometric features indicating the location of facial characteristic points (mouth, eyes, chin, etc.) or appearance features representing the facial textures \citep{Zeng2009, Valstar2015}. In \citep{Pantic2006}, a set of facial points was used as geometric feature to recognize AUs in frontal-view face images. In \citep{Bartlett2006, Guo2005}, appearance-based methods, such as Gabor wavelets or eigenfaces were applied to classify facial expressions or AUs. Some other work used both geometric and appearance features. In \citep{Ringeval2015}, appearance features were extracted using local Gabor binary patterns and geometric features were extracted based on 49 facial landmarks. Similarly, \citep{Benitez-Quiroz2016} used second-order statistics of facial landmarks (i.e., distances and angles between landmark points) for geometric features and Gabor filters for appearance features.

However, most 2D feature-based methods are only suitable for the analysis of frontal-view face, which means only a small range of head movement is allowed. Fewer works of facial expression analysis have been done based on 3D face models. In \citep{Zeng2006}, a 3D face tracker was used to handle the arbitrary behavior of the person in the natural setting. Both geometry and appearance features were extracted based on the 3D face model. In \citep{Cohn2004}, authors focused on recognizing two of the most important facial actions in brow (brow raising and brow lowering) measured in spontaneous facial behavior with non-frontal pose, moderate out-of-plane head motion, and occlusion. A cylindrical head model was applied to estimate head movements in \citep{Xiao2002}.


To date, research on fusion of facial expression and physiological data in order to improve performance of emotion recognition algorithms is continuing to attract the attention of academia and industry alike.
In \citep{Koelstra2013}, multimodal approaches based on both feature-level and decision-level fusion were applied to analyze facial expressions and EEG signals for generation of affective tags. In feature-level fusion, the authors simply stacked all the feature vectors together. In decision-level fusion, they first classified each modality individually and then combined the classifier outputs in a linear fashion. In \citep{Liao2006}, authors focused on recognizing only two different affective states, stress and fatigue. They applied a decision-level fusion approach to recognize these two affective states from multiple modalities including physical appearance (e.g. facial expression, head movement), physiological measures (e.g. EEG, ECG), behavioral data (e.g. mouse movement, type speed), and user performance (e.g. response time). In \citep{Miao2018}, authors proposed to apply recurrence network analysis to quantify the dynamics and to extract nonlinear features from EEG signals for classifying affective states. Based on existing knowledge and methods, the fusion of multimodal person-specific data in levels prior to the emotion experience inference is largely unexplored \citep{Zeng2006}. It is worth mentioning that data fusion in earlier stages, before decision, can lead to capturing the higher order  information presents in the multimodal data, as well as the dynamical coupling between different input modalities.


\section{Materials \& Methods}
\rating
\subsection{Dataset}
In  this  work, we used a multimodal dataset collected by the Psychology Department of Northeastern University, which contains both facial video recording and synchronous physiological data including ECG, EDA, and respiration signals. These data were obtained from 12 consenting participants during two data collection phases: (Phase I) each participant described their two most positive and their two most negative emotional experiences, and (Phase II) each participant watched their own 4 recorded videos as stimuli.
In both phased, facial videos were recorded by a frontal camera at 25 frames per second. The three physiological signals were sampled at 1000Hz using BioLab v.3.0.13 (Mindware Technologies; Gahanna, OH) via a BioNex 8-Slot chassis (Model50-3711-08).

The recorded videos in Phase I (that formed our video-based stimuli) were played back to the same individual who recorded them (i.e., participants watched themselves), such that each participant viewed video-based stimuli in Phase II. Participants provided continuous ratings of their affective feelings while watching their video-based stimuli, using a rating dial (ranging from unpleasant to neutral to pleasant). Continuous ratings were obtained since this method is non-disruptive (are not requiring stopping of the stimulus), allows online ratings that previous research suggests are less subject to recall bias, and can provide idiographic data at a high temporal resolution \citep{ruef2007continuous}. These self-ratings, as shown in \figref{rating}, reveal dynamics across the video-based stimulus segments in the degree of positive or negative affect. 

\subsection{Facial Landmark Localization and Tracking}
In order to track relative movements of the facial landmarks solely generated by the facial muscles in the videos, we developed a robust tracking approach, in which the head movement is also tracked and decoupled from the facial landmark movements. We first employed a state-of-the-art 2D facial alignment algorithm presented in \citep{kazemi2014one} to automatically localize 68 landmarks for each frame of the face video. Then, a 3D face model based on \citep{kittler20163d} is used to estimate the depth information from the 2D frames and thus to achieve 3D landmark tracking.

\landmarkFramework

\subsubsection{2D Facial Landmark Localization}
A cascade of trained regressors is utilized to localize the facial landmarks for each video frame as described in \citep{kazemi2014one} (see \figref{landmarkFramework}). To train each regressor, the gradient tree boosting algorithm is used with a sum of square error loss \citep{john2010elements}. Assume we have training dataset \{($I_1,S_1$), ... , ($I_n,S_n$)\}, where each $I_i$  is a face image and $S_i$ is its shape vector. We set an initial shape estimate $\widehat{S}_{i}^{(0)}$ for every face image. In each regression tree, the regression function $r_t$ is learned using the gradient tree boosting algorithm, and then the estimation of every shape is updated as:

\begin{equation}
\widehat{S}_{i}^{(t+1)} = \widehat{S}_{i}^{(t)} + r_t(I_i,\widehat{S}_{i}^{(t)})
\end{equation}

The initial shape $\widehat{S}_{i}^{(0)}$ for each frame is simply chosen as the mean shape of the training dataset centered and scaled according to the bounding box of the full face, detected with the histogram of oriented gradients (HOG) features. In each level of cascade, estimated landmarks are refined by adding residuals produced by the previous regression tree. Note that all frames in video are normalized to have the same Euclidean distance of pixels between the middle of the two eyes. Hence the landmark movements would be comparable across a given individual. \figref{MohsenLandmark} shows the result of 68 landmark detection on a video frame captured by a laptop webcam.

\MohsenLandmark

\subsubsection{3D Facial Landmark Tracking}
Our facial landmarks tracking algorithm needs to remain invariant across head movement including its translation, scaling (getting closer or further from camera), and rotations (i.e. roll, yaw, pitch). To eliminate the interference of head movement, we first extract the depth information of each face pixels from 2D video frames using the 3D morphable face model described in \citep{kittler20163d}. This model consists of a principal component analysis (PCA) model of face shapes, which could be used for reconstructing a 3D face from a single 2D image. The PCA model consists of a set of principal components $V = [\upsilon_1,...,\upsilon_K]$, the mean value of all the facial meshes $\overline{\upsilon}$, and their standard deviation $\sigma_k$. The shape of a novel face is then generated with:

\begin{equation}
S_i=\overline{\upsilon} +\sum_{k=1}^{K} \alpha_k \sigma_k \upsilon_k,
\end{equation}
where $K$ is the number of principal components and $\alpha_k$'s are the representation of $S_i$ in the coordinates of the PCA shape space. The 3D face shapes were then reconstructed by fitting 68 detected landmarks to a PCA shape model. For the purpose of model fitting, the gold standard algorithm of Hartley and Zisserman \citep{hartley2003multiple} were implemented, which finds a least squares approximation of an affine camera matrix given 2D-3D point pairs. An example result of the landmark fitting is shown in \figref{YuModel}.

\YuModel

Using the 3D geometric transformation matrix, every frame could be transformed into frontal face, enabling us to track and compare the movement of the facial landmarks throughout the video. The landmarks detection algorithm is not accurate for some frames with large head rotation angles or poor lighting. We address this issue by applying landmark geometric constraints.


\subsubsection{Facial Expression Features}
Rather than few prototypic facial expressions, such as happiness, anger, surprise, and fear, it is shown that the dynamics and temporal combination of facial action units (AUs) may provide more reliable and specific quantification of the expressive movements of the human face during emotion \citep{tian2001recognizing}. 
Guided by the work done by Y. Tian \citep{tian2001recognizing}, we reduced the facial landmarks feature dimensions from 2$\times$68 to 12 features described as follows: (1-2) left and right eyebrow y-values (corresponding to AUs 1, 2 and 5); (3) inner corners differences of eyebrows (corresponding to AU 4); (4) horizontal distance of the the two corners of lips (corresponding to AUs 12 and 20); (5) vertical distance of the two lips (corresponding to AUs 25, 26 and 27); (6) average vertical positions of the two corners of the lips (corresponding to AU 15); (7-9) head rigid displacement in X, Y, and Z direction, respectively;  (10-12) head rigid rotation in roll, pitch, and yaw direction, respectively.
It is noted that combining the facial landmarks into AUs leads to the reduction of the stochastically distributed noise in landmark positioning as well as resolving the issue of subject-specific and camera-specific variations. 


Moreover, comparing to other open source tools that can directly detect AUs, our method can provide more information related to the facial movements for two reasons. Firstly, most AU detection algorithms could only deal with frontal-view face, while our 3D landmark tracking method could also extract head movement directly from the video. Secondly, our method provides continues measures of facial landmarks instead of only 6 discrete numbers representing AUs and its intensities. 













\subsection{Peripheral Physiological Signal Processing}

\subsubsection{Signal Preprocessing}
To eliminate or reduce the noise and artifacts carried by the physiological signal measurements, signal-specific filtering is required prior to applying any feature extraction algorithm. An elliptic bandpass filter with the cut-off frequencies of 5Hz and 45Hz was applied on the ECG signals. This cut-off frequency range was selected based on the power spectral density analysis of the ECG signals and the elliptic filter type was selected to ensure the amplitude of the peak points on the signal were not significantly suppressed by the filter \citep{Chavan2006elliptic}. With the similar investigation, a low-pass filter with cutoff frequency of 1Hz was selected as the optimal choice and was applied on the EDA signals \citep{deluca2010emg}. As for the respiration signal, we applied a Butterworth lowpass filter with a cutoff frequency of 20Hz.



\subsubsection{Psychologically-Inspired Physiological Features}
The dimensionality reduction of the peripheral physiological time series signals is done by employing a series of signal-specific algorithms to extract informative physiological features from each signal \citep{NabianMohsen2017Abpt}. Physiological signals including ECG, EDA, and respiration were processed using the biosignal processing MATLAB toolbox accessible at \citep{AClabtoolbox}.

ECG signals contain rich information relevant to  human health,  sleep quality, and emotional states \citep{Ra2013ECG}. For the ECG signals, the features are based upon the detected QRS points on the signal \citep{1985real} and after successfully detecting these points, relevant physiological features can be computed (see \tblref{BioFeatures}).


\tabBioFeatures

The EDA signal is composed of two types of activity, tonic and phasic. The slowly varying base signal is the tonic aspect, and is also called skin conductance level (SCL). The faster-changing part is called phasic activity or skin conductance response (SCR). SCRs are related to more acute exterior stimuli or non-specific activation \citep{gamboa2005electrodermal}. Many important features for this purpose are extracted from SCRs. The occurrence of the SCR is detected by finding two consecutive zero-crossings, from negative to positive and positive to negative of the bartletted differentiated EDA signal. Most of EDA features listed in \tblref{BioFeatures} are based on the detection of SCRs.
It is noted that for given windows of EDA in which no SCR was found, feature values were set to 0. 

Features were also extracted from respiration signal as its pattern may vary in distinct affective states. Regular respiration is linked to relaxation, while fast and shallow breathing might correspond to more aroused emotions, such as acute anxiety and emotional tension \citep{Koelstra2013}. Important features extracted from respiration signal are provided in \tblref{BioFeatures}.

\subsubsection{Windowing and Overlapping Sizes for Feature Extraction}

Multimodal data fusion using a recurrent network requires all feature vectors to be the same length. Therefore, values for window sizing and overlapping were carefully selected for different facial expression signals as well as physiological signals to achieve same size feature vectors in all of the modalities as well as to reasonably capture the temporal dynamics of the signals. Please note that the larger window size is chosen for the signal with slower changing rate to make sure appropriate features are correctly extracted. To align different modalities in time, we applied different overlaps to force corresponding windows of different modalities to have the same starting point in time. Windowing and overlapping sizes for each signal modality are provided in \tblref{FeatureExtraction}.

\tabFeatureExtraction

\graphFlow
\subsection{Multimodal Data Fusion using Recurrent Network}

Here, we demonstrate a method introduced earlier by \citep{Zou2015} to build an inter-system network model that represents the joint contribution between different response modalities of a person under an affective experience context. In our study, the nodes in the network represent features extracted from multimodal data (facial expressions and physiological responses) and edges are defined based on the directional coupling between modalities. The network construction consists of the following steps: (1) time-delayed embedding for reconstructing phase space trajectory \citep{Takens1981}; (2) recurrence plot (RP) construction \citep{Eckmann1987}; (3) extension of RP to multiple systems (i.e. modalities) to obtain a joint RP (JRP) \citep{Romano_2004}; and (4) extraction of complex network metrics to assess the inter-system dynamical coupling. The general framework of the proposed fusion approach applied on our multimodal dataset is shown in \figref{graphFlow}.


\subsubsection{Recurrence Plot for Single Modality}
Introduced by Eckmann in 1987 \citep{Eckmann1987}, RP is a visualization to represent the temporal dependency relationships between all states in a time series data using a binary, squared matrix \citep{Eckmann1987}. Suppose the state of system (or modality) $X$ at time \textit{i} and {j} is represented by $\textbf{x}_i, \textbf{x}_j$, recurrence can be recorded by the binary function as:
\begin{equation}
\textbf{R}_{i,j}^{X} = \Theta (\epsilon_X -{||\textbf{x}_i - \textbf{x}_j||_1}), \textbf{x}_i \in \mathbb{R}^m, \textit{i}, \textit{j} = 1,...N,
\end{equation}
where $\Theta$ is a Heaviside function and the RP puts a point at coordinates $(i,j)$ if ${R}_{i,j}^{X} = 1$, any time the state trajectory gets sufficiently close (within the system threshold $\epsilon_X$) to a point it has been previously. 

\subsubsection{Extension to Multimodality and Investigation of Coupling}
Although the original method was developed for a single time series, later variations of RP included consideration of multivariate time series from different aspects. To best capture the dynamic coupling between multiple modalities, we adopted JRP since it represents when a recurrence occurs simultaneously in two or more time series \citep{Romano_2004}. Suppose we have modalities $X$ and $Y$, for which the individual RPs can be obtained. In general, JRP is obtained by the product of multiple systems:
\begin{equation}
\textbf{JR}_{i,j}^{X,Y} = \Theta (\epsilon_X -{||\textbf{x}_i - \textbf{x}_j||_1}) \Theta (\epsilon_Y -{||\textbf{y}_i - \textbf{y}_j||_1}).
\end{equation}


\subsubsection{Complex Network-based Feature Extraction}
We enabled the usage of complex network analysis by converting the JRP matrix to the adjacency matrix for a network, which serves as a graphical representation of the temporal neighborhood relations between system states across entire time series. This network, referred to as recurrence networks (RN), is expressed in the following formula:
\begin{equation}
\textbf{A}_{i, j}(\varepsilon) = \textbf{JR}_{i, j}(\varepsilon) - \textbf{I},
\end{equation}
where $\textbf{A}_{i, j} (\varepsilon)$ is the adjacency matrix, $\textbf{JR}_{i, j} (\varepsilon)$ is the JRP (a binary matrix), and $\textbf{I}(T)$ is an identity matrix for removing the elements on the main diagonal line that creates self-loops in the network.

In our study, RN is used to describe the dynamical behaviors (or patterns) of a person under an affective experience context. Further, to characterize the network features, a set of metrics are computed for quantitative assessment of the network topology, and further provide information about coupling dynamics in a different view. The network measures we computed include two general classes: (1) global measures: transitivity, global efficiency, and out-strength/in-strength correlation, and (2) local measures: in-strength/out-strength, local efficiency, edge/node betweenness centrality, diversity, and clustering coefficients. The global measures are related to the topological structures of the entire network, while the local measures are related to the attribute of individual nodes. These metrics are widely used for describing the connectivity patterns in complex systems; the computational details are included in \citep{rubinov2010complex}.

\subsection{Inference Models for Affective Experience Prediction}


\subsubsection{Prediction based on Emotional Video Stimulus Content}
We performed a binary across-individual classification on video stimuli with positive or negative contents. Since each subject experienced two positive and two negative video stimuli, the class distribution was balanced. 
First, we extracted the signal-specific features from the multimodal input signals collected from each subject while he/she was watching a video stimulus. To fuse these features, we then constructed their JRPs, as well as the network metrics containing global and local measures. We used a support vector machine (SVM) with a linear kernel function as the inference model and the stimuli context prediction accuracy was defined as the percentage of stimuli correctly classified based on their positive or negative contents. We performed a per-subject leave-one-out cross validation method for the model evaluation.

\subsubsection{Prediction based on Affective Self-Rating Scales}
Here, we trained an inference model to predict individualized affective experiences of participants while watching a video stimulus, rather than classifying the content of the stimulus. This inference model accounts for individual differences in responding to emotional stimuli, by assuming that the multimodal facial and physiological data collected during an affective experience is more directly linked to the person's internal affective state rather than the content of the stimulus as positive or negative.

The same signal-specific features are fused by JRP for predicting the valence self-rating of the person. The original range of valence rating was [0, 50]. We applied a Min-Max normalization to scale the valence ratings to [0, 1] for each subject. Since the actual rating range of different subjects could differ a lot, normalized values allow the comparison across subjects in a way that eliminates the effects of certain gross influences. For each video stimulus, participants provided continuous ratings of their positive/negative affect. We trained our inference model to predict the median value of continuous ratings, since self-ratings might become dramatically high or low at the end of the video. Instead of predicting the mean value of continuous ratings, we assumed the median value is a more robust representation of the subject's valenced experience during the whole video stimulus.

We employed support vector regressor (SVR) with a ridge penalty as the self-rating regression model. The same leave-one-out cross-validation approach is applied as in the previous classification task. Root mean square error (RMSE) and mean absolute error (MAE) of the predicted valence self-rating scores are employed here to indicate the performance of our trained regression model.

\section{Experimental Results}

\tabclassificationResult
\subsection{Emotional Video Stimulus Content Prediction Results}
We performed a per-subject leave-one-out cross validation, where the classifier is trained on a total of 44 trials from 11 participants, and tested on the 4 videos from the remaining one participant. For each participant, classification accuracy and F1 score were used to evaluate the performance of our stimulus content classification model. \tblref{classificationResult} gives the classification results over each modality and video stimulus content. As a baseline, we also gave the expected values of the random classifier. According to the significance test with the resulted p-values in \tblref{classificationResult}, the obtained classification accuracies of the trained model are not significantly more accurate than the random classifier. 
Overall, the result shows that video content ratings might not be predictable from the recorded data. One explanation for the low prediction accuracy of our trained model could be the fact that the data was not recorded from the participants while they were verbally explaining their positive or negative experiences, but rather we used the facial and physiological signal recordings of the participants while they were watching back their own recorded videos. In other words, all the recorded signals are not directly related to the video-based stimulus contents. In the following section, we instead predict the self-ratings, which may be more closely related to the recorded multimodal signals.


\tabregressionResult
\subsection{Affective Self-Rating Scale Prediction Results}

\tblref{regressionResult} provides the results for our regression model to predict self-ratings of the subjective experience of emotion trained on the multimodal fusion data. We reported the RMSE and MAE, as the two widely used metrics for the accuracy evaluation of continuous variable estimation. Both of these values are negatively-oriented scores, in which the lower values represent higher accuracy performance. As a baseline, we also gave the expected values of a random regressor, which were found by randomly generating the value that are drawn from a normal distribution of the training data.

As evident in the \tblref{regressionResult}, the prediction accuracy measures obtained from the models trained on either single modality and multimodal fusion data are higher than the prediction accuracy of the random classifier, except for the models trained solely on head movement features. Moreover, the highest accuracy was achieved  with our proposed network-based multimodal fusion method. Note that the physiological signals are also well ranked in terms of performance for rating prediction: respiration performs second best on valence and EDA performs third best. This implies that alongside the facial video data, the physiological signals provide informative complementary descriptions of the affective experience.

In addition to our multimodal fusion method based on the recurrence network, another basic fusion method was also employed and evaluated. In the basic method, each modality of facial (i.e. landmarks and head movement) or physiological signals (i.e. ECG, EDA and Respiration) is weighed equally for the fusion. The results of facial and physiological features fused by the basic fusion method were also given in \tblref{regressionResult}. It is observed that the prediction accuracy from fusion of facial signals outperforms the prediction accuracy based on single modalities. However, the model based on fusion of the physiological signals only outperformed the model trained on the ECG features and had lower performance than models trained on EDA and respiration data.

\section{Conclusion}
In this paper, we presented a multimodal approach that analyzes both facial expressions and peripheral physiological signals concurrently to identify and decode the individualized affective experiences of participants watching a series of emotional video-based stimuli. We developed a robust 3D face tracking approach, in which head movement is also independently tracked and decoupled from the facial landmark movements. Signal-specific features were then extracted from both facial and physiological signals (ECG, EDA, and respiration). We applied recurrence network for multimodal data fusion and complex network-based features were extracted from the fusion of different modalities. Finally, we validated our approach using a multimodal dataset consists of (i) facial videos and (ii) several peripheral physiological signals, synchronously recorded from 12 participants while watching 4 emotion-eliciting video stimuli. The experimental results for binary classification of video-based stimuli with positive vs. negative content showed that video content ratings might not be predictable from the recorded data from the participants when they are watching them.
One potential explanation is that people might recall their feelings while watching themselves talking about past experiences, but they may not feel the same at that moment. However, the recorded signal modalities were shown to be reliable predictors of the self-rating of their affective experience at the moment of watching the videos. Our feature-level fusion approach based on the recurrence network demonstrated to improve upon single modality results, suggesting the these modalities contain complementary information in accounting for person's affective experiences. 
\bibliographystyle{named}
\small
\bibliography{paper}

\end{document}